\RequirePackage{fix-cm}
\documentclass[twocolumn]{svjour3}          
\smartqed  
\usepackage{graphicx}
\usepackage{color}
\begin{document}

\title{A Recursive Framework for Expression Recognition: From Web Images to Deep Models to Game Dataset
\thanks{This work is supported by the National Science Foundation through Award EFRI -1137172, and VentureWell (formerly NCIIA) through Award 10087-12. }
}


\author{Wei Li  \and Christina Tsangouri    \and Farnaz Abtahi  \and  Zhigang Zhu 
}



\institute{Wei Li \at
              Dept. of Electrical Engineering, CUNY City College  \\
              \email{wli3@ccny.cuny.edu}           
           \and
           Christina Tsangouri \at
           Dept. of Computer Engineering, CUNY  City College \\
           \email{ctsango000@citymail.cuny.edu}
           \and
           Farnaz Abtahi \at
           Dept. of Computer Science, CUNY Graduate Center \\
           \email{fabtahi@gradcenter.cuny.edu}
           \and
           Zhigang Zhu \at
           Dept. of Computer Science, CUNY City College and Graduate Center \\
           \email{zhu@cs.ccny.cuny.edu}
}	     

\date{Received: date / Accepted: date}

\maketitle

\begin{abstract}
In this paper, we propose a recursive framework to recognize facial expressions from images in real scenes. Unlike traditional approaches that typically focus on developing and refining algorithms for improving recognition performance on an existing dataset, we integrate three important components in a recursive manner: facial dataset generation, facial expression recognition model building, and interactive interfaces for testing and new data collection. To start with, we first create a candid-images-for-facial-expression (CIFE) dataset. We then apply a convolutional neural network (CNN) to CIFE and build a CNN model for web image expression classification. In order to increase the expression recognition accuracy, we also fine-tune the CNN model and thus obtain a better CNN facial expression recognition model. Based on the fine-tuned CNN model, we design a facial expression game engine and collect a new and more balanced dataset, GaMo. The images of this dataset are collected from the different expressions our game users make when playing the game. Finally, we evaluate the GaMo and CIFE datasets and  show that our recursive framework can help build a better facial expression model for dealing with real scene facial expression tasks.  
  
\end{abstract}

\section{Introduction}

Detecting people's  facial expressions has been an interesting research topic for more than 20 years. Facial expressions play an important role in many applications such as advertising, social interaction and assistive technology. Research in facial expression recognition mainly includes three parts: datasets, algorithms and real world interaction applications. In our approach, to improve the performance of facial expression recognition in real scenes, we propose a framework for integrating dataset construction, algorithm design and interaction implementation.

\begin{figure}
  \includegraphics[width=0.45\textwidth]{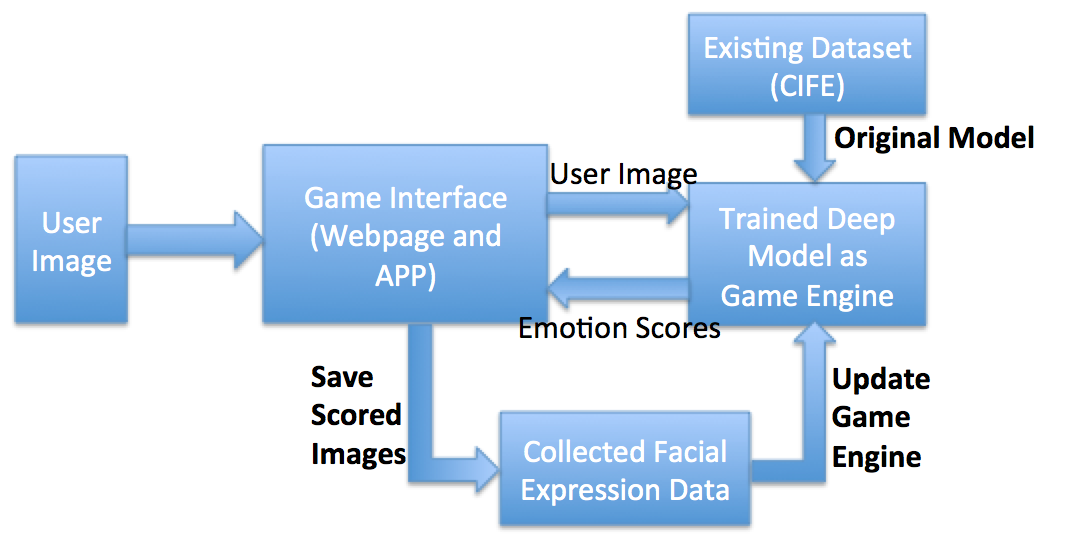}
\caption{The proposed recursive framework}
\label{fig0}       
\end{figure}

Even though facial expression research should include three integrated parts, most of the previous work mostly focused  on one of the components. Consequently, algorithms or models designed for one or several datasets do not work well when dealing with real scene problems. With this in mind, we propose a recursive updating approach. Starting from a deep learning model trained from facial images collected from the Web, a facial expression game is designed for collecting new and more balanced data. Then the newly collected data are used to update the training model. The framework is illustrated in Figure \ref{fig0}. We start with a dataset with Candid Images for Facial Expression (CIFE) to build an initial facial expression model, which is then served as the game engine for a facial expression game, then when users play the game, facial images of the users  are classified as different facial expressions by the model and automatically collected. This leads to a new and balanced dataset, named GaMo (standing for Game-based eMotion), which is used to update our facial expression model. 

\subsection{Contributions of the paper}
This is an extended work of our previous research in web-based data collection CIFE ~\cite{li2015deep}, CIFE data enahancement and AlexNet model fine-tuning ~\cite{li2015emotiw} and game interface designs for collecting new data (GaMo) \color {black} \cite{li2016cvpr} \color {black}.
In this paper we have the following new contributions: 

1) A recursive framework is proposed, which can recursively generate new data and update the deep learning model to have better performance in real scene facial expression recognition. 

2) A deeper CNN model is used by fine-tuning the 19-layer CNN structure proposed by the Visual Geometry Group (VGG)\cite{vgg2014} - thus the fine-tuned VGG network, and performance comparison showed that it outperformed the results we generated using our initial CNN model and the fine-tuned AlexNet model we reported in our previous work. 

3) We detailed the design and evaluation of the game interface, which is only controlled by human facial expressions, and automatically collecting expression images while the players are playing the game. The new GaMo dataset is also analyzed leading to insights and new ideas for more balanced data collection with our recursive framework in the future.

4) The performance of emotion recognition based on the two facial expression datasets is compared and analyzed: CIFE and GaMo, and their more balanced subsets using the new VGG model: CIFE is a web candid imaged based facial expression dataset; GaMo is an in Game-based eMotion dataset collected when our users played our facial expression game.


\subsection{Organization of the paper}
The rest of the paper organized as follows: After the introduction in Section 1, related work is reviewed in Section 2. Section 3 discusses how the CIFE dataset is collected and applied to build a CNN based  facial expression model. Details of how we design and implement  the game interface is described in Section 4. We evaluated our framework by comparing GaMo and CIFE datasets in building facial expression recognition models in Section 5. And finally, we concluded our work in Section 6.
 
\section{Related Work}

We already mentioned that current facial expression research mainly includes three major components: datasets, algorithms/models, and applications. Here we would like to give a review of each of the three components. 

\subsection{Datasets}
Among the many datasets that have been provided by researchers for recognizing expressions from images, there are mainly two kinds of datasets. Datasets belonging to the first category are captured in laboratory. These include  CK+, MMI and DISFA dataset ~\cite{mavadati2013disfa,kanade2000comprehensive,cohn2007observer,lucey2010extended}. Usually subjects are invited to their labs and sit in lighting and position constrained environments. Good results can be achieved on these datasets but in real life scenarios, it's always hard to have good performance.
Datasets in the second category are collected from existing media and social networks, such as Kaggle and EmotiW ~\cite{li2015deep,li2015emotiw,pantic2005web}. Using web search engines,  one can easily obtain thousands of images but the datasets are usually not balanced. EmotiW is a video clip dataset for an expression recognition challenge, and the video samples are from Hollywood movies where the actors show different expressions. For the datasets collected from existing media, some of the expressions like Happy or Sad are easier to obtain, but for some expressions like Disgust or Fear, it's hard to find enough samples. 

\subsection{Algorithms and models}
Although the existing datasets are generally not balanced, many interesting and promising approaches have been proposed for expression detection. Most existing facial expression 	recognition methods have focused on recognizing expressions of frontal faces, such as the images in CK+ \cite{lucey2010extended}. Shan, et al \cite{shan2009facial} have proposed a LBP-based feature extractor combined with an SVM for classification. In the method proposed by Xiao, et al \cite{xiao2011facial}, instead of training one model for all expressions, separate models have been trained for each expression, which improve the overall performance. Wang, et al \cite{wang2013capturing} modeled facial expressions as complex activities that consist of temporally overlapping sequences of face events. Then, an Interval Temporal Bayesian Network (ITBN) was used to capture the complex temporal information.  Karan, et al ~\cite{sikkaexemplar} proposed a HMM-based approach to make use of consecutive frame information to achieve better expression recognition accuracy from video.

In the last few years, 
deep learning methods have been successfully used for face recognition and verification \cite{taigman2014deepface,sun2014deep}. Deep learning approaches are also used in many expression detection applications. Liu, et al \cite{liu2014facial} proposed a Boosted Deep Belief Network to perform feature learning, feature selection and classifier construction for expression recognition. Different DBN models for unsupervised feature learning in audio-visual expression recognition have been compared in the work done by Kim, et al \cite{kim2013deep}. Our early work  \cite{li2015deep} used CNNs on images collected from the Web. To prove the effectiveness of CNNs, we compared CNN-based facial expression performance on CK+ to the state of the art methods. Multimodal deep learning approaches have been applied to facial expression recognitions tasks. An example is Jung, et al's work ~\cite{jung2015joint} in which facial landmarks based shape information and image based appearance information are learned through a combined CNN network. The results show that deep learning based multimodal features act better than individual modalities or the use of traditional learning approaches. Automatically learned features have also been used for multimodal facial emotion recognition on video clips ~\cite{li2015emotiw}.


\subsection{Interactive applications}
In the generation of ImageNet dataset ~\cite{deng2009imagenet}, Amazon Mechanical Turk (AMT) is used to label all the training images. Workers are hired online and can remotely work on labeling the dataset. The ImageNet is a large scale dataset that aims to label 50 million images for object classification and without the help of online workers, the labeling would not be feasible. This inspired us to develop the idea of involving people in the data collection process through an online framework, preferable using games. There have been some efforts in using games to attract people to perform some image classification work. Luis, et al~\cite{von2004labeling} designed an interactive system that attracted people to label images, Mourao et al ~\cite{mourao2013competitive} developed a facial engaging algorithm as the controller to play their Novoexpressions game, and a player engagement dataset was obtained and the relationship between the players' facial engagement and game scores were analyzed. But their goal is not to collect data. Expression games have also been used to entertain children with Autism Spectrum Disorder (ASD) and to help them perform facial expressions by mirroring their expressions to some cartoon characters ~\cite{deriso2012expression}. Our online framework not only makes use of  online crowdsourcing through games, but also has much lower cost than AMT. And since the numbers of various facial expressions can be controlled by the design of the games, the dataset can be guaranteed to be balanced. 

\section{CIFE: A Dataset with Web Images}


Since we would like to develop facial expression approaches that can be used in real world scenes, we need to train and test the models on non-posed images, or candid images. Therefore we collect a Candid Image Facial Expression (CIFE) Dataset. We note that most of the facial expression images on the Web are randomly posed, and most of the expressions are natural. Therefore we use web crawling techniques to acquire candid expression images from the Web, and create our candid image facial expression dataset CIFE. 

\subsection{CIFE data collection}
\begin{figure}
  \includegraphics[width=0.4\textwidth]{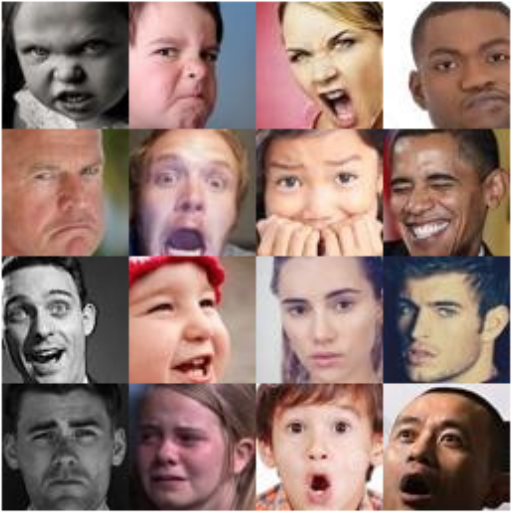}
\caption{Images from CIFE}
\label{fig1}       
\end{figure}

As we have mentioned, we define seven types of expressions: Happy, Anger, Disgust, Sad, Surprise, Fear and Neutral. Using  related key words to the each of the 7 expressions in addition to the name of the expression (e.g.,  joy, cheer, smile for Happiness), we have collected a large number of images that belong to each of the seven expressions. \color {black} We have used most of the image search engines, including Google, Baidu and Flickr\color{black}. In our initial CIFE dataset ~\cite{li2015deep}, the number of samples of different expressions were: Anger (1785), Disgust (266), Fear (781), Happiness (3636), Neutral (644), Sadness(2485) and Surprise(997). The images are from the web and most of them are not posed.  However, the number of samples in different classes was highly  unbalanced. Therefore, we have added some images to classes with fewer samples (for example Disgust and Fear) to balance the class sizes  ~\cite{li2015emotiw}. At the end, we obtained 14,756 images for 7 classes  (after some manual post-filtering by humans). The total number for each facial expression  in our revised CIFE dataset is listed in Table \ref{tab1}. This is the dataset we use in this paper. In  ~\cite{li2015emotiw}, the CNN model was one of the modules for video expression recoginition, but here we focus on facial expression recognition in single images. Figure \ref{fig1} shows a few typical examples of faces with various poses.
\begin{table}
\caption{Sample numbers of the seven facial expressions in CIFE (Ang, Dis, Fea, Hap, Neu, Sad, Sur represent angry, disgust, fear, happy, sad, surprise, respectively).}
\label{tab1}       
\begin{tabular}{|c|c|c|c|c|c|c|c|}
\hline
 Expr &Ang &Dis&Fea&Hap&Neu &Sad &Sur \\
\hline Nums&1905&975&1381&3636&2381&2485&1993 \\
\hline
\end{tabular}
\end{table}

\subsection{CIFE data augmentation}
\begin{figure}
  \includegraphics[width=0.4\textwidth]{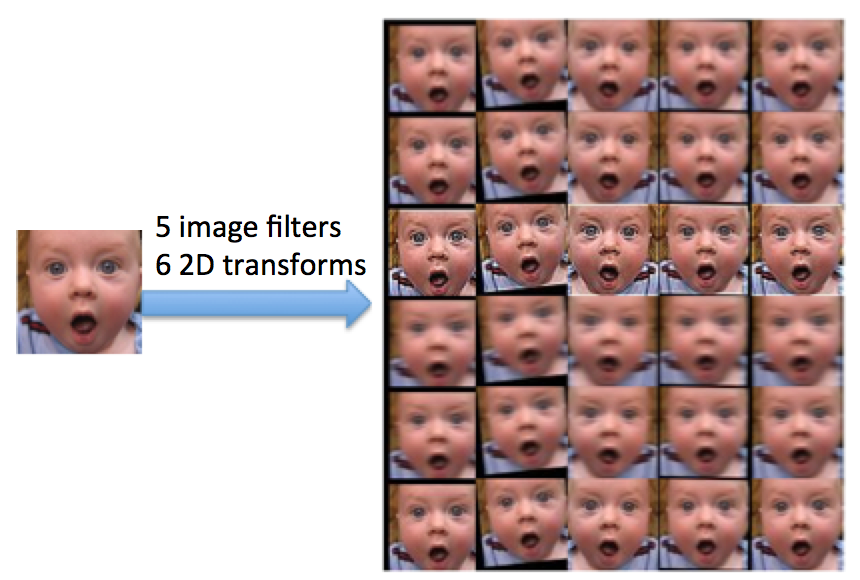}
\caption{CIFE data augmentation}
\label{fig_aug}       
\end{figure}

Deep learning  with CNNs always requires a very large number of training images in order to train a large number of parameters of the network for obtaining good classification results. Even though our CIFE dataset has 14,756 images for 7 classes, it is still insufficient for training a deep CNN model. So before training the CNN model, we need to augment the dataset with various transformations to generate various small changes in appearances and poses. We applied five image appearance filters and six affine transform matrices. The five filters are disk, average, gaussian, unsharp and motion filters, and the six affine transforms are formalized  by adding slight geometric transformations to the identity matrix, including a horizontal mirror image. \color {black} Figure \ref{fig_aug} \color {black} shows an example of the facial image augmentation.  By doing this augmentation, for each original image in the dataset, we  can generate 30 (=5x6) samples, therefore the number of possible training samples would increase from 10330 be 309900, which is sufficient for training the deep learning model. 

\section{Fine-tuned CNN Models}


 After data augmentation, we now have 309900 training images, and the model will be tested on 4,424 original testing images (30\% of 14,756). Our goal is to classify all the images into 7 facial expression groups. To achieve our goal we design an CNN structure. In the following, we will describe our initial CNN model, the fine-tuning of two CNN structures - ALexNet  \cite{alex2012} and VGG \cite{vgg2014} , and report the comparison of their performance.

\subsection { The initial CNN model}
Our initial CNN model structure includes one input layer (the original image), three convolutional layers, and an output layer. This structure was arrived by trial and error with many experimental tests. The input color image size is 64x64, and the number of the output is 7. We set the convolutional filters size to be 3x3. We then varied the number of layers and the number of filter for each layer. After many rounds of tests, we finally arrived at the 'best' structure with 3 convolutional  layers, and the filter numbers for each layer to be 32, 32, 64, respectively. For each of the three convolutional layers, we add a 2:1 pooling layer to make the training data less redundant. The input 64x64 RGB image is then recognized as one of the  7 labeled classes. With this structure, we can easily know the numbers of the parameters to be around 184,000. Compared to the number of training images (309,900), the structure setting is also appropriate. Finally, we achieved a 65.2\% accuracy on our test data. Even though the accuracy was not very high, the CNN-based facial expression showed its obvious performance advantage over tradition approaches such as the results using support vector machines (SVMs), 62.3\% with the LBP Feature and 59.7\% with the SIFT feature. Details of the results with the traditional approaches can be found in our previous work at ~\cite{li2015deep}. We want to note here that we reported a much higher classification accuracy (81.5\%) using a similar CNN model. That was because the highly unbalanced numbers of samples in the initial CIFE dataset used in ~\cite{li2015deep}: there the recognition rates for disgust and fear classes were very low, for both the original dataset and the revised dataset, and hence adding new samples decreased the overall recognition statistics. We suspect that the reason for the low performance was that the three-layer structure is unable to learn the features deeply enough.   
%

 \begin{figure}
  \includegraphics[width=0.50\textwidth] {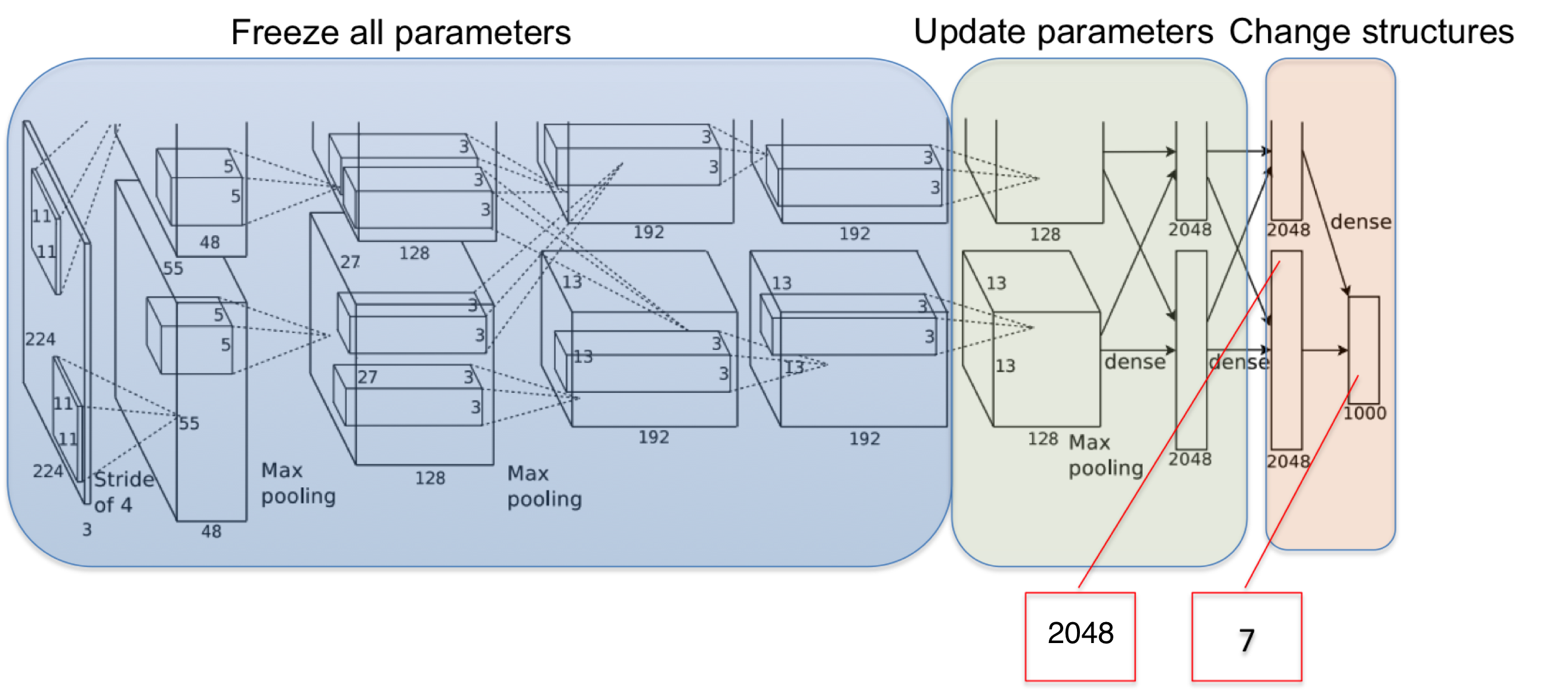}
\caption{Fine-tuning AlexNet structure for facial expression recognition}
\label{fig2}       
\end{figure}

\subsection { Fine-tuning AlexNet}

To further improve the performance of facial expression recognition using CNN, we noted that learned general classification models can be used for specific classification problems \cite{frcnn}. Since some existing learned models are deeply trained on large scale datasets, image features thus learned can be better features for recognition of other classification tasks. Therefore we are curious to find out if this can help improve facial expression recognition. To try out our idea, we did experiments by fine-tuning AlexNet \cite{alex2012} and VGG  \cite{vgg2014} structures.
 
In the AlexNet structure, there are 1 input layer, 5 convolutional layers and 3 fully connected layers, leading to 60 million parameters in total. Our first guess was that training the AlexNet on our CIFE dataset would results in better classification accuracy. The only problem was the need for a  larger number of images, as the ImageNet requires millions of images during training. 

Therefore, we instead propose a CNN fine-tuning method to train a deeper model based on AlexNet. The rationale is that although our task is different from the ImageNet, which focuses on object classification, similar low level filters could be used in expression recognition. Based on this hypothesis, we can use the AlexNet and utilize our relatively 'small' dataset to update and fine-tune parts of its parameters for adapting it to expression recognition. 

As shown in Figure \ref{fig2}, the parameters of the convolutional layers 1 through 4 are not changed. Our new CIFE dataset is used to update the parameters of the convolutional layer 5 and the first fully connect layer, without changing their structures. In the original AlexNet, the number of units of the second fully connected layer and the third layer are 4096 and 1000 (classes) respectively. Since the number of classes in our dataset is just 7, we needed to change the structure in these two layers. We reduced the number of neurons in the penultimate layer to 2048, and the third fully connected layer to 7.  The classification accuracy by using this model is 73.5\% on the revised CIFE dataset, which shows that the fine-tuning leads to a much better performance than our first attempt of using a three-layer CNN structure, a 8.3\% improvement. This was the model we used in collecting the GaMo emotion dataset, when only this model was available.
With this decent accuracy, a system that uses such a facial engine can lead to a good chance to obtain the right prediction in human computer facial interaction to encourage users to play the interaction game, which will be described in the next section. 

\subsection { Fine-tuning VGG}
Compared to AlexNet, VGG is a much deeper network. After the GaMo data collection, we also investigated if using a  fine-tuned VGG model can improve the facial expression recognition performance. We first tested the fine-tuned VGG model on the revised CIFE dataset for comparing with the results with the fine-tuned AlexNet model.  There are 19 learning layers in total, \color {black} with 138 million parameters \color{black}. VGG layers have some similar structure as AlexNet. They both have convolutional parts and fully connected parts. For each convolutional layer in AlexNet, VGG replaces it with 2-4 convolutional layers. Deeper networks lead to better representation of the input images: in the ImageNet challenge, VGG yielded a 6.8\% top-5 error compared to AlexNet's 16.4\%. We applied a similar fine-tuning approach to the VGG net as we did to AlexNet. By fintuning on existing VGG model with the revised CIFE dataset, we finally achieved a 76.3\% accuracy, which is a 2.8\% improvement over the fine-tuned AlexNet, and 11.1\% over our initial CNN model. Therefore in Section 6, we will show results using the fine-tuned VGG structure for emotion recognition with CIFE and GaMo datasets and their sub-sets.

Through finetuning the ImageNet models, we obtained improved facial expression recognition results. It indicates that the models for general image classification can share convolutional filters with specific purpose image classification tasks, such as facial expression recognition. In this way the fine-tuning leads to more robust models for non-posed image facial expression prediction. 

\section{GaMo: Game Based Interface for Balanced Expression Data}
We already showed that deep learning can lead to high accuracy facial expression recognition. While the candid images are most likely randomly posed, but they are still different from images from real scenario interaction. More "real" data would be facial expression images of people collected without any constraints. If people can show this kind of "real" facial expression, we can use our facial expression model to select corresponding images and construct a real scene emotion dataset. The selected images may be useful for building a real scene expression model. For this purpose, we decided to design a game interface that invites people to show their facial expressions, willingly, while playing a game.

  \begin{figure}
  \includegraphics[width=0.4\textwidth] {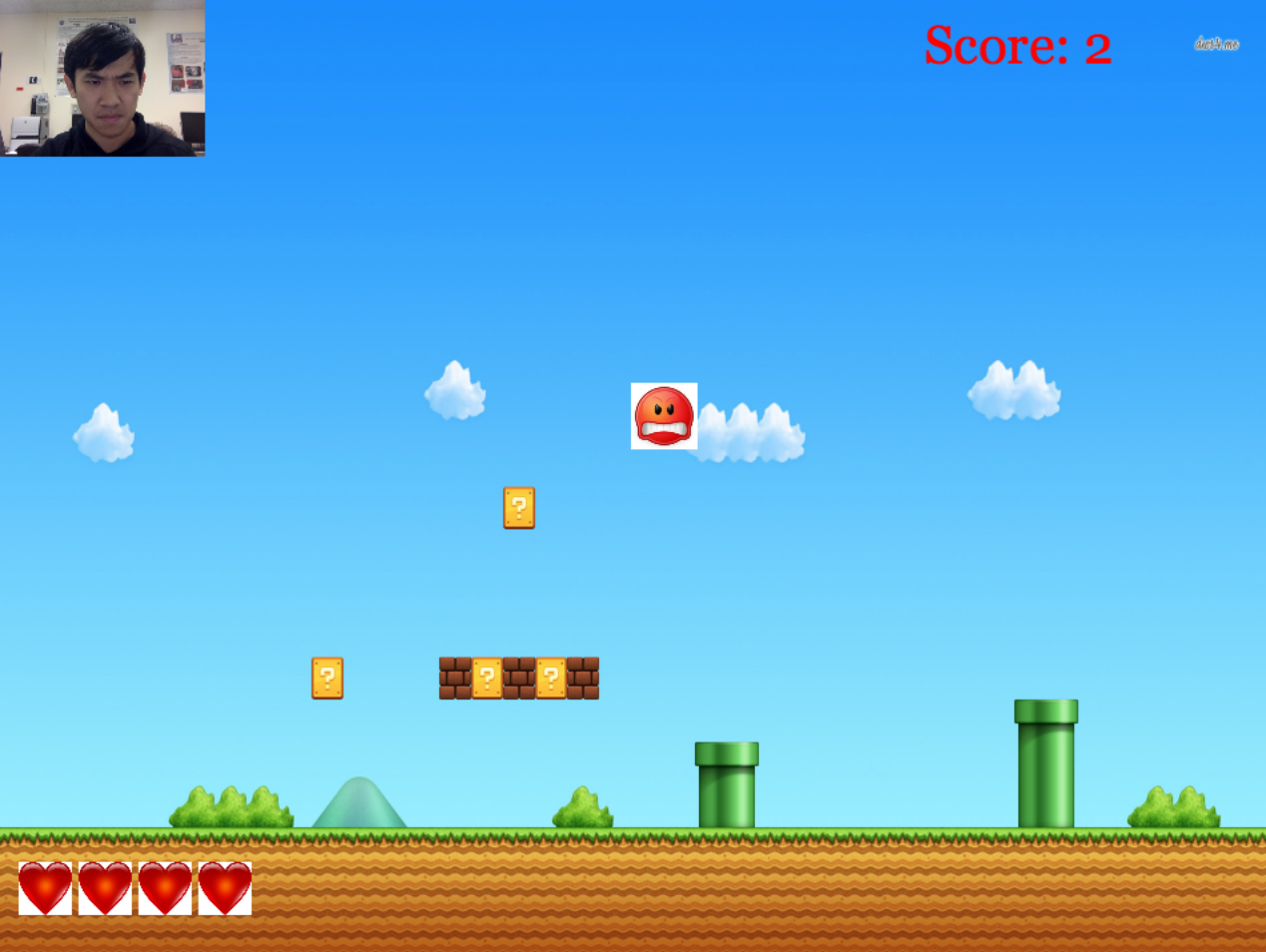}
\caption{Design of the facial expression game scene}
\label{fig3}       
\end{figure}

\subsection{Game design}
\begin{figure}
  \includegraphics[width=0.5\textwidth] {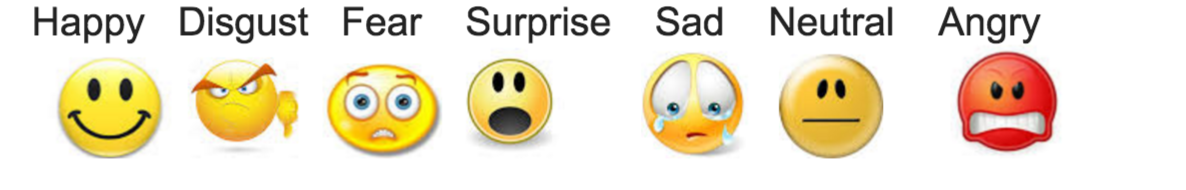}
\caption{Facial icons representing the 7 basic expressions}
\label{fig4}       
\end{figure}

Since we would like game users to show real facial expressions yet remain engaged throughout the game, we had to design the expression game in a way that it is straightforward and interesting as much as possible. After performing research on the popular web games, we found that the Tower Defense games is the style that fits our task the best. The logic of the Tower Defense games is always very simple: the player needs to build a defense system against the intruders. In our application, we would like the user to act as a defender against an "expression target". An example of the basic game scene is shown in Figure \ref{fig3}. The live video of the user is shown on the top-left corner of the screen so he can always see his expression. A randomly picked expression target as shown in Figure \ref{fig4} (a facial expression icon) will enter the screen as a bomb dropped from above and the user has to protect the village by making the bomb disappear before it reaches the ground. Some sound effects are also added to make user more engaged. The bomb would disappear if the user makes a facial expression that correctly matches the displayed expression (the bomb), as judged by the CNN-based expression detector using the fine-tuned AlexNet model which was the available model to us when we collected the new data.
The score as shown on the top-right of the screen will be increased. 

\subsection{General version and customized version}
Here we would like to provide some technical details of our general game design. The expression game web interface accesses the camera on the user's machine and displays the video on the top-left corner of the screen. Then the game interface captures images of the user's face,  and then sends face images to our server. The CNN model we trained by fine-tuning AlexNet analyzes each image and generates a probability vector for the seven expressions and sends it back to the game webpage. The reason for processing face images at a server is due to the high computational requirements  by the CNN model.  After the probability vector of the seven expressions is sent back to the game interface, it compares this feedback with the expression target ID that has been displayed on the screen as a bomb and informs the server to save the image if it matches the icon. Since the target facial expression that the user needs to make is defined by our system, we not only know the label and have high confidence that facial expression's label is correct, but also can make the dataset more balanced based on our needs. Our game can be accessed via this test page \footnote { http://emo.vistawearables.com/modeltests/new (Firefox tested and recommended)}.

The frame rates of typical webcameras are usually 20-60 Hz. We do not need such high frame rates of image capturing for two main reasons: 1) From the computational resource's point of view, the server will have a huge workload if we run the expression recognition on every single image since the CNN computation is time-consuming  \color {black} with hundreds of millions of parameters \color{black}.  2) There is no need to know the user's expression frame by frame. Giving the user some time delay to prepare their expression may actually result in better image quality and as a consequence, a much better dataset. For these two reasons, we design the game in such a way that it only sends one image per second to the server. It takes only around 200 ms for the server to generate the results for every single image, which makes the game run very smoothly. We also set the number of initial game lives to be five and generate the expression targets randomly, with equal probabilities to all the seven expressions, which theoretically result in a balanced dataset. 

The game is implemented using Javascript and HTML. The backend is hosted using Ruby on Rails. During the game, expression icons will drop from the top of the screen, and the user's face is also shown in the left top corner of the screen for the user to check her/his facial expression. If the user is able to match the expression before the icon hits the ground, she/he will receive 1 point. The score will change as the user gains or loses points. When all the game lives are used, a "Game Over" sign will be shown up, together with the total score gained by the user,  and then a "Replay" button.

After making the game available to a small group and collecting data from several users who tried it, we realized that the collected dataset is not ideal, specifically for two main reasons: 1) Sometimes it is hard for users to correctly imitate the exact expressions by just looking to the icons; 2) Our expression detector sometimes is not able to correctly determine whether the subject is making the right face or not. This makes it hard for the players to achieve high scores and as a result, the collected data becomes imbalanced.

We were able to provide two solutions to solve the above problems. 
One is to change the expression recognition to an expression verification task. This makes the classification task much easier. Since we know the "ground truth" or the target label for an icon being displayed in the game, we only need to check if the probability of this specific expression reached a predefined threshold. Each expression needs to have its own threshold since some expressions are harder to mimic through facial expressions and have higher variety among different users. This will help the users achieve higher scores and also include a broad range of correctly labeled facial expressions for each expression in the dataset. 

Another solution is to create an individual model for each player based on the CNN extracted features. The user will be compared with her/his individual expression templates instead of the general CNN model. The Deepface work \cite{taigman2014deepface}  has proved that the CNNs is not only able to directly perform image classification, but can also extract robust features from the images. Thus we extract the features from the CNN for each individual user and then these features are saved as  templates for that specific user. This makes the game customized for each player and the user can gain higher scores and is encouraged to play more.

  \begin{figure}
  \includegraphics[width=0.5\textwidth] {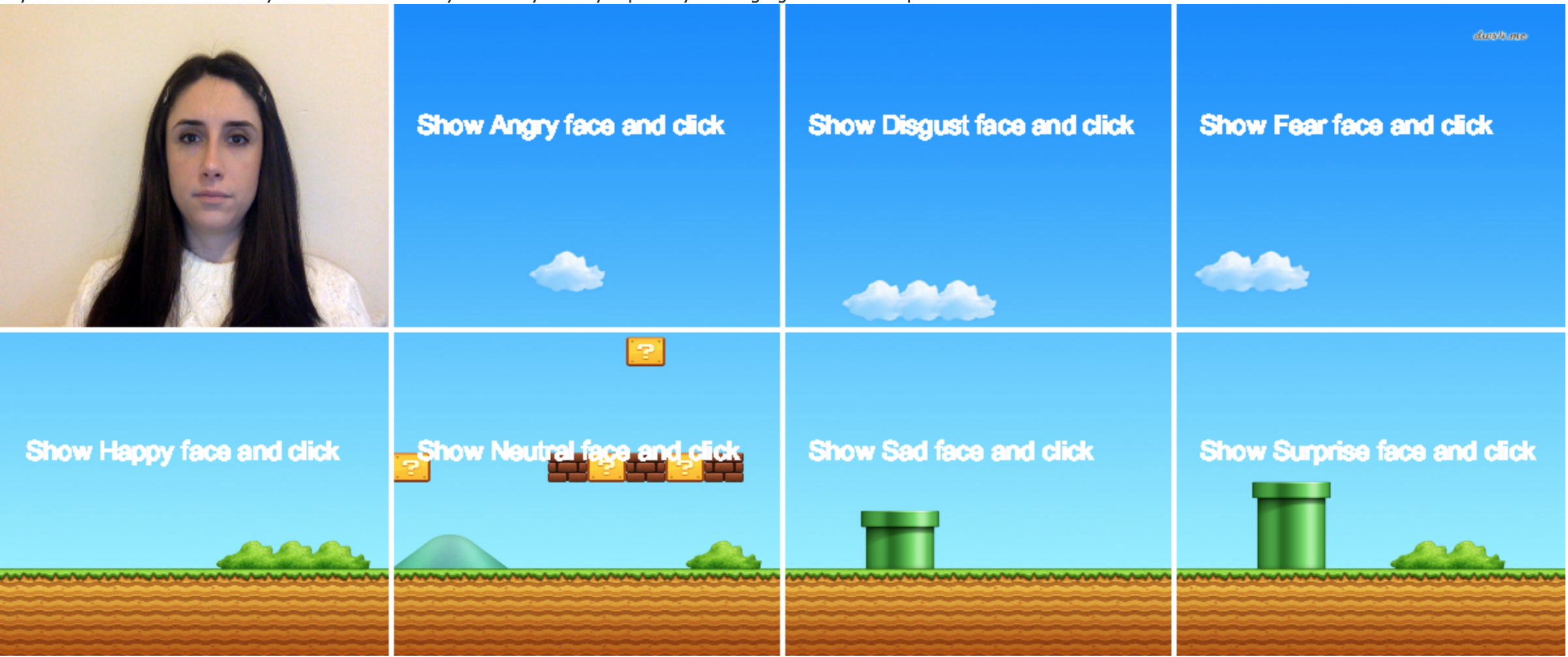}
\caption{Registration page for building customized facial templates: initial page. The first image is the live webcam view, and the rest seven are places for showing the templates of the seven expressions. }
\label{fig5}       
\end{figure}

\begin{figure}
  \includegraphics[width=0.5\textwidth] {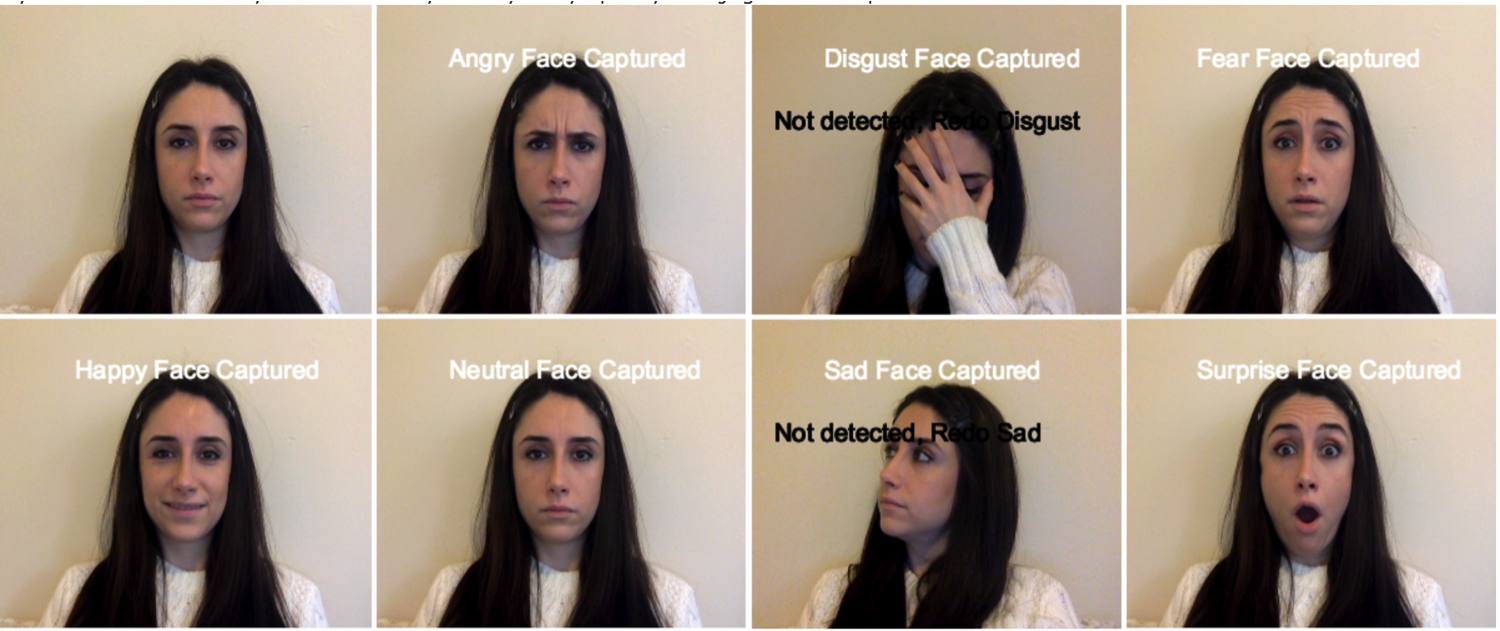}
\caption{Faces of the seven expression after the user click "Send All". In this example, two of the images are not qualified so the user has to recapture these two images.}
\label{fig6}       
\end{figure}

As a part of the solutions proposed above, we designed a user registration page as shown in Figure \ref{fig5}. The registration page is divided into 8 subareas. The first subarea shows the current video stream. The other seven subareas display the seven registered expression templates. To save each template image, the user can click on the corresponding subarea while imitating the correct facial expression. This process can be repeated several times until the user is happy with the saved image. Once all seven expressions are registered, the user can click the "Send All" button to send the expression templates to the server, where the system will detect the face area in the images and use the CNN model to extract expression features for the user. If the face cannot be detected, an error message will be sent back to the user, and she/he is then asked to recapture the image for the specific expression that has caused the error, as shown in Figure \ref{fig6}. The register page can be accessed here \footnote {http://emo.vistawearables.com/usergames/new (Firefox tested and recommended)}.

When all the template features are saved in the server, the user will be directed to the customized game scene. While the game is being played, the server will extract features for the image that is being sent at the moment and compare these features to the saved emotion templates. We use L2 distance to select the nearest result and send it back as the detected emotion. Since the features are robust and the user is always compared with her/his own model, the user will potentially achieve a higher score. We call this version of the game the "customized version", as opposed to the previous "general version".

\subsection{GaMo: the New Dataset}
Within one month of the release of the two game test versions to the college students of our department, more than a hundred users played the general version and 74 users tried the customized version. All the users that we collected data from have signed the consent form of our IRB approval. We obtained 15455 images in total during this time period and generated the GaMo (game based expression) Dataset. Compared to some deep learning datasets, the size is still not big enough, but our game can run at any time, so we can obtain a much bigger dataset when the game reaches more people. The dataset is available by contacting the authors. 

 One concern for our dataset might be the use of a trained model to get more expression data: Will this recognition/ verification model just take expression data that are similar to our existing data samples and make the dataset less diverse? We believe that by collecting more data from more people, the dataset will be much more diverse. While the general model can contribute to the overall data diversity, the customized version may only collect users specific data similar to their templates. Although  for each individual, a type of expression tends to be very similar, by assembling all the people together, the data is still diverse. And the most important thing is, deep learning can learn the common features of each expression well if we can feed all possible data to it. 

We would like to note here that no manual cleanups for the images and labels have been done; all the images are used in our evaluation in Section 4. By randomly checking the dataset, we have not found any labels that are very off the real expressions. The distribution of the dataset is shown in Table \ref{tab2}. Compared to the CIFE dataset, GaMo is more balanced, which hopefully will result in a much more reliable facial expression detector. In conclusion, the data collection is automatic, of high quality and more balanced. We will evaluate our dataset in the next section.

\section{Evaluation}
Based on our proposed framework, we applied deep learning and finetuning to web collected images CIFE and obtain a facial expression recognition model - the fine-tuned AlexNet, then we used the model to host the facial expression game to collect the new GaMo data. We hope the GaMo dataset can be used to recursively finetune our CNN models. To prove this, we first need show that the GaMo data can actually improve the real scene facial expression recognition.

To determine the usefulness of the GaMo dataset, we performed the following experiments. First, we trained a new CNN model with GaMo by finetuning the previous AlexNet model that has been used as our game engine, which was trained on CIFE . To compare GaMo with CIFE, we ran both a self evaluation and a cross evaluation with the two CNN models: the GaMo CNN model and the CIFE CNN model. 

In our earlier work \cite {li2016cvpr}, we have shown that the model trained with the more balanced GaMo dataset produced more robust results, especially for those classes that were underrepresented in the CIFE dataset. Further the GaMo model can be applied to the CIFE dataset with a decent performance, but not the other way around. 

In order to see if these observations are consistent with more complicated and better performed models, we then used the fine-tuned deeper VGG models - the best models among the ones we have developed and used. The VGG models are trained with the CIFE and GaMo datasets, respectively, and performed the same experiments as in our earlier work.

We noted that due to the game engine we used which was based on the unbalanced CIFE dataset, we still had fewer samples in some categories, thus the new GaMo dataset was not completely balanced. Therefore we also ran an experiment to see if we just using more balanced sub-sets  from both CIFE and GaMo will have large changes to the recognition results. Finally, we designed a small user study to find out if the dataset can actually improve the game engine and game experience. For this purpose, the users played the general version of the game hosted by the two new CNN models.

\subsection{Comparison of CIFE and GaMo}

\begin{table}
\centering
\caption{Comparison of expression sample numbers in CIFE and GaMo}
\label{tab2}
\begin{tabular}{|c|c|c|} \hline
Dataset& CIFE & GaMo \\ \hline
Angry& 1905& 1945\\
Disgust& 975& 1838\\
Fear& 1381& 1586\\
Happy& 3636& 3185\\
Neutral& 2381& 2741\\
Sad& 2485& 1898\\
Surprise& 1993& 2262\\

\hline\end{tabular}

\end{table}

Table \ref{tab2} show the statistics of GaMo and CIFE datasets. For the CIFE dataset, as we mentioned before, the images are collected by  searching from web engines using key words. We also went through the dataset to remove all the images that are not meaningful facial expressions. To some extent, the numbers of samples in the seven emotion categories reflect the distribution of facial images numbers online. We can clearly see the imbalance of the sample numbers for different facial expression categories, and it's hard for us to balance it since if we use the minimal number 975 for Disgust, the numbers of samples would be too small.  We can also see that the sample numbers of each facial expression from GaMo are more balanced. Although we noticed some of the expression numbers are also smaller than others like Fear and Disgust, it's easy for us to make it more balanced. When we design the game, we make all expressions show up at the same probability, but due to different ability of the expression prediction using the game engine trained with the unbalanced CIFE dataset, the GaMo dataset is not completely balanced. The good news is that, since we already have known the different accuracy in predicting each facial expression, in the future data collection with our recursive framework, we can change the show-up probabilities of of facial expression targets to control the final data distribution. This is our ongoing work.
 
\subsection{Comparison of CNN models with CIFE and GaMo }

To compare the two models, we test the overall accuracy in recognizing all the seven expressions (\color {black} the average accuracy in Table \ref {tab3} \color {black}) as well as the accuracy of each individual expression within its own sub-dataset (Angry, Disgust, Fear, Happy, Neutral, Sad, and Surprise), as listed also in Table \ref{tab3}, and in \color {black} Table \ref{tab4} to Table \ref{tab7} \color {black}. This would give us a good sense on the usefulness of GaMo dataset. Furthermore, to compare the performance of the two CNN models based on the VGG structure, we perform a cross evaluation: the model trained on CIFE is tested on images from GaMo and vice versa. 

\begin{table}
\centering
\caption{Average accuracies of self and cross evaluation of CIFE and GaMo models}
\begin{tabular}{|c|c|c|c|c|} \hline
     &CIFE& GaMo& CIFE cross& GaMo cross\\ \hline
Average& 0.76&  0.75&  0.31&  0.64\\
\hline\end{tabular}
\label{tab3}
\end{table}

\begin{table}
\centering
\caption{Self evaluation confusion matrix of CIFE}
\label{tab4}
\begin{tabular}{|c|c|c|c|c|c|c|c|} \hline\smallskip
     &Ang& Dis& Fea& Hap& Neu& Sad & Sur\\ \hline
Ang&  0.81&  0.03&  0.02&  0.01& 0.03& 0.06& 0.03\\
Dis&  0.07&  0.53&  0.06&  0.03& 0.19& 0.03& 0.06\\
Fea&  0.04&  0.02&  0.62&  0.02& 0.04& 0.07& 0.2\\    
Hap&  0.02&  0.02&  0.001&  0.85& 0.02& 0.05 & 0.02\\
Neu&  0.05&  0.09&  0.02&  0.02&  0.70& 0.04& 0.08\\
Sad&  0.07&  0.01&  0.01&  0.03& 0.04& 0.82& 0.01\\
Sur&  0.01& 0.01& 0.08& 0.02& 0.07& 0.01& 0.78 \\
\hline\end{tabular}
\end{table}

\begin{table}
\centering
\caption{Self evaluation confusion matrix of GaMo}
\label{tab5}
\begin{tabular}{|c|c|c|c|c|c|c|c|} \hline\smallskip
     &Ang& Dis& Fea& Hap& Neu& Sad & Sur\\ \hline
Ang&  0.62&  0.07&  0.02&  0.04& 0.11& 0.08& 0.04\\
Dis&  0.06&  0.69&  0.03&  0.06& 0.05& 0.08& 0.01\\
Fea&  0.02&  0.05&  0.62&  0.05& 0.11& 0.02& 0.1\\    
Hap&  0.02&  0.02&  0.01&  0.81& 0.04& 0.05 & 0.02\\
Neu&  0.01&  0.02&  0.02&  0.03&  0.85& 0.02& 0.02\\
Sad&  0.02&  0.05&  0.02&  0.05& 0.04& 0.77& 0.02\\
Sur&  0.02& 0.02& 0.06& 0.04& 0.04& 0.01& 0.79 \\
\hline\end{tabular}
\end{table}

\begin{table}
\centering
\caption{Cross evaluation confusion matrix of CIFE}
\label{tab6}
\begin{tabular}{|c|c|c|c|c|c|c|c|} \hline\smallskip
     &Ang& Dis& Fea& Hap& Neu& Sad & Sur\\ \hline
Ang&  0.11&  0.06&  0.03&  0.02& 0.48& 0.0& 0.29\\
Dis&  0.02&  0.18&  0.01&  0.02& 0.50& 0.01& 0.25\\
Fea&  0.01&  0.05&  0.04&  0.02& 0.26& 0.0& 0.6\\    
Hap&  0.01&  0.01&  0.01&  0.02& 0.34& 0.01 & 0.53\\
Neu&  0.01&  0.02&  0.01&  0.01&  0.85& 0.0& 0.11\\
Sad&  0.002&  0.06&  0.008&  0.008& 0.71& 0.01& 0.19\\
Sur&  0.004& 0.01& 0.03& 0.017& 0.13& 0.04& 0.79 \\
\hline\end{tabular}
\end{table}

\begin{table}
\centering
\caption{Cross evaluation confusion matrix of GaMo}
\label{tab7}
\begin{tabular}{|c|c|c|c|c|c|c|c|} \hline\smallskip
     &Ang& Dis& Fea& Hap& Neu& Sad & Sur\\ \hline
Ang&  0.71&  0.02&  0.01&  0.01& 0.01& 0.13& 0.03\\
Dis&  0.13&  0.28&  0.01&  0.16& 0.01& 0.36& 0.02\\
Fea&  0.01&  0.02&  0.44&  0.14& 0.02& 0.11& 0.18\\    
Hap&  0.02&  0.01&  0.001&  0.91& 0.01& 0.04 & 0.01\\
Neu&  0.12&  0.05&  0.02&  0.14&  0.24& 0.39& 0.03\\
Sad&  0.07&  0.01&  0.01&  0.08& 0.02& 0.82& 0.01\\
Sur&  0.04& 0.01& 0.11& 0.17& 0.02& 0.05& 0.58 \\
\hline\end{tabular}
\end{table}

The confusion matrices of these four experiments are listed in \color {black} Table \ref{tab4} to Table \ref{tab7} \color {black}. 
Looking into the self evaluation results, we can see that the model trained on GaMo has a much more balanced distribution on expression classification on all the seven expressions. Even though the average performance of the CNN model on the CIFE is  slightly higher than that on GaMo, the numbers are misleading since the higher average accuracy of the CIFE-trained CNN model is due to the much larger numbers of samples in both Happy and Sad classes, which apparently also have much higher accuracy than others. In comparison, the performance in recognizing Disgust and Fear is much higher using GaMo than using CIFE. 

\begin{figure*}
  \includegraphics[width=0.9\textwidth] {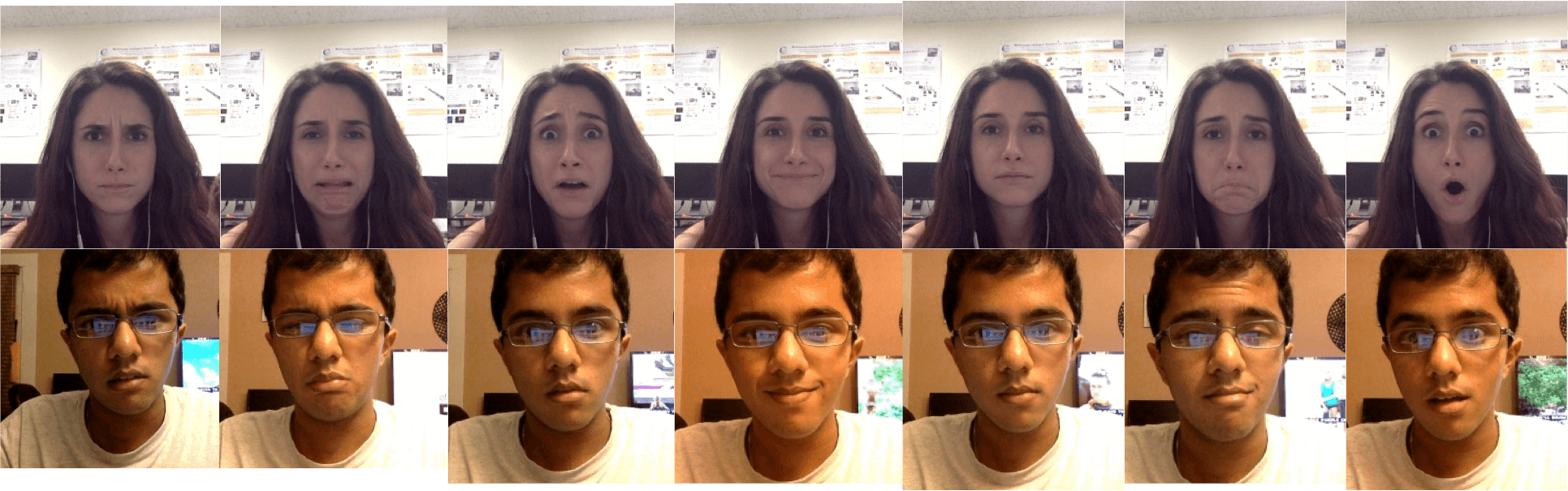}
\caption{Comparison of individual template images of two users from GaMo}
\vskip -6pt
\label{fig10}
\end{figure*}

The results of the cross dataset tests are even more interesting. The model trained on CIFE has a very low performance when tested on the GaMo dataset, the confusion matrix shows that many images are classed to neutral. We have observed that the difference between the images is significant among the two datasets. Our observations indicate that the expressions in the CIFE dataset tend to be more exaggerated and thus easier to be identified, as it are shown in Figure 6, while the GaMo dataset is more realistic to real life, as it is obtained from ordinary users with a high amount of varieties in imitating facial expressions while playing the game. As an example, Figure \ref{fig10} shows two users who played the game. The first player shows more explicit expressions while the second player's expressions tend to be more implicit. This makes it hard for the model trained on CIFE to classify the images from GaMo. The CIFE model almost completely fails in recognizing Angry, Disgust and Happy in GaMo. We believe the reason is that these three expressions in the CIFE dataset, whether they have fewer or more samples, are much more highly exaggerated than those in the GaMo dataset. On the other hand, when the model trained on GaMo is cross-tested on CIFE, the performance is surprisingly good, even though the performance cannot beat that on the self-test. The reason is that the model is further fine-tuned on a larger, more inclusive and more balanced dataset. The GaMo model does reasonably well on all the three expressions failed by the CIFE model.  In addition, if subtle expressions (as in the GaMo dataset) can be recognized, the exaggerated ones (as in CIFE) are not difficult to detect. As an example, the Happy faces in CIFE can be much more easily recognized (with a 91\% accuracy) using the GaMo model.  

Here we also want to note that the performance using the VGG structure is much better than using the AlexNet; interested readers please compare the results in Table \ref{tab3} with the results in our previous work \cite {li2016cvpr}. Nevertheless, the performance comparison observations between the CIFE and GaMo datasets are consistent from the AlexNet to VGG structure.

\begin{figure*}
  \includegraphics[width=0.9\textwidth] {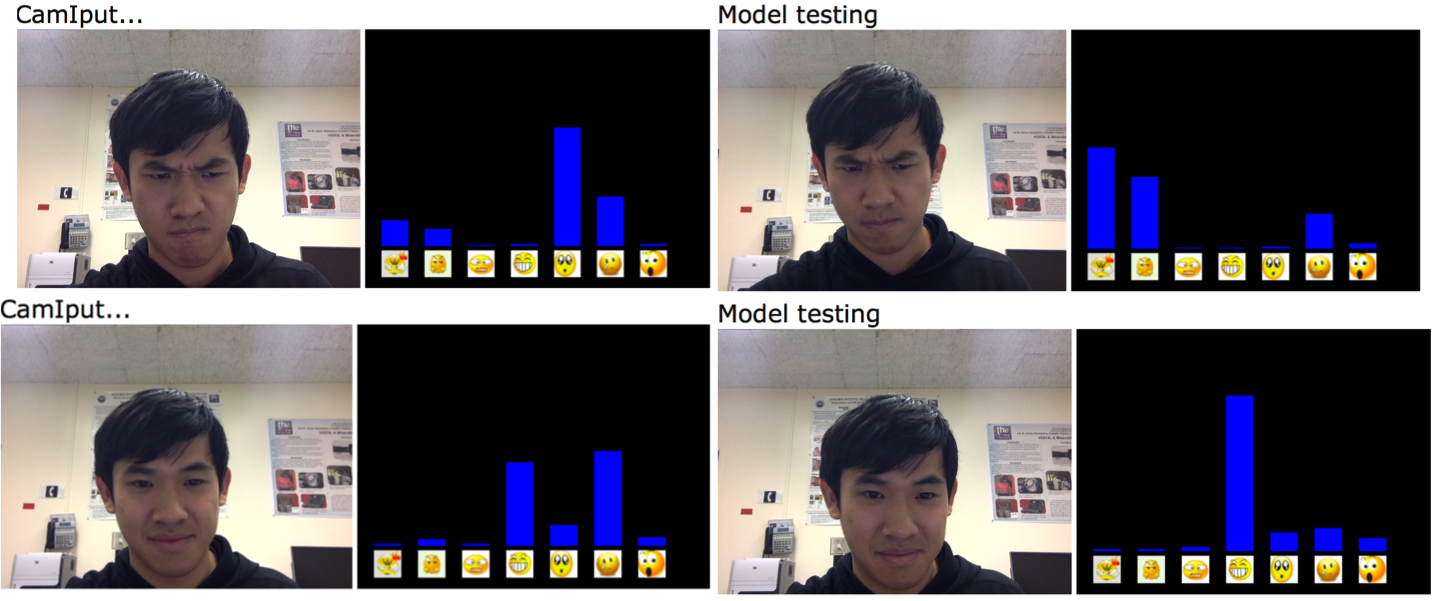}
\caption{Subtle facial expression recognition by CIFE and GaMo models (left two are from the CIFE model and right two are from the GaMo model). Each histogram shows the probability distribution of the seven emotions for each facial image. The order of the expressions is Angry, Disgust, Fear, Happy, Neutral and Surprise. For some subtle expressions, only the GaMo model works well.}
\label{fig7}       
\end{figure*}

\subsection{Comparison with more ''balanced" sub-datasets}
In most facial expression datasets, the sample numbers for different expression are imbalanced. The training process favors the class which has more samples to get higher accuracy. But this will weaken the model's ability to recognize the facial expression with less samples. In reality, this will not be a good interactive experience if the facial expression model is unable to recognize some less frequent facial expressions. So if we want to build a model that can recognize all the facial expressions with equal accuracy, the best way is to create a training dataset with similar samples.  
In this case, for dataset CIFE, we will only have 4781=683x7  (683 is the number of Disgust expression samples in the CIFE train set, 70\% of the total) images in total, which may not be sufficient for training a well-performed deep learning model. We use the subset of the CIFE dataset, which is balanced  and run the same deep learning training as the full CIFE. We augment the 683 images of each facial expression and then finetune the VGG model. The final prediction result on CIFE is shown in Table \ref{tab8}. As predicted, by comparing Table \ref {tab4} and Table \ref {tab8}, overall the performance is lower than using the full CIFE dataset, \color {black}except the least frequent expressions: Disgust and Fear:  The average recognition rate drops by 9\%. \color {black} But for the GaMo dataset, we still have over 10K images  in the subset of balanced data, and the training data set has \color {black} over 7770 images (1586x70\%x7) \color {black}.  The performance result using the balanced GaMo subset is shown in Table \ref{tab9}. Compared to the result in Table \ref{tab5}, the overall performance improves \color {black} by 6.4\%. As a matter of fact, the recognition rates for all the categories increase; those with lower numbers of samples in the original GaMo datasets (Fear, Disgust and Sad) increase significantly, by more than 10\%. \color {black}

By comparing the two approaches in collecting facial expression images: searching from the Web, and harvesting from game users, we have some important notes.
First,  it's almost impossible for us to get more facial images for CIFE as we already have searched most of the image search engines in order to obtain high quality images. While for GaMo, as long as our game is running, we can have more and more balanced expression data. 
Second, even for the current version of GaMo, we retrained the deep learning model with the balanced subset of the GaMo dataset and by testing on the same original GaMo testing data, we see the performance has increased significantly.  

\begin{table}
\centering
\caption{Self evaluation confusion matrix of sub-balanced CIFE}
\label{tab8}
\begin{tabular}{|c|c|c|c|c|c|c|c|} \hline\smallskip
     &Ang& Dis& Fea& Hap& Neu& Sad & Sur\\ \hline
Ang&  0.71&  0.04&  0.05&  0.06& 0.04& 0.19& 0.03\\
Dis&  0.06&  0.55&  0.06&  0.03& 0.19& 0.03& 0.06\\
Fea&  0.04&  0.02&  0.64&  0.02& 0.04& 0.05& 0.15\\    
Hap&  0.05&  0.02&  0.04&  0.73& 0.03& 0.07 & 0.03\\
Neu&  0.05&  0.14&  0.05&  0.02&  0.60& 0.04& 0.08\\
Sad&  0.11&  0.04&  0.04&  0.04& 0.05& 0.68& 0.01\\
Sur&  0.03& 0.03& 0.08& 0.02& 0.07& 0.01& 0.67 \\
\hline\end{tabular}
\end{table}

\begin{table}
\centering
\caption{Self evaluation confusion matrix of sub-balanced GaMo}
\label{tab9}
\begin{tabular}{|c|c|c|c|c|c|c|c|} \hline\smallskip
     &Ang& Dis& Fea& Hap& Neu& Sad & Sur\\ \hline
Ang&  0.68&  0.08&  0.08&  0.02& 0.06& 0.03& 0.02\\
Dis&  0.02&  0.83&  0.03&  0.03& 0.02& 0.03& 0.03\\
Fea&  0.01&  0.01&  0.85&  0.02& 0.04& 0.05& 0.05\\    
Hap&  0.02&  0.03&  0.02&  0.84& 0.03& 0.03 & 0.02\\
Neu&  0.02&  0.03&  0.04&  0.02&  0.86& 0.04& 0.02\\
Sad&  0.02&  0.03&  0.03&  0.02& 0.02& 0.87& 0.01\\
Sur&  0.02& 0.02& 0.06& 0.02& 0.02& 0.01& 0.86 \\
\hline\end{tabular}
\end{table}

In the balanced CIFE subset, due to fewer training data than the full CIFE, the performance for Angry and Happy dropped dramatically but the accuracy for Disgust and Fear doesn't improve much. While for the GaMo dataset, since each facial expression still has more than 1586 images, the balanced subset of GaMo is still a good dataset for training. The balanced GaMo produced a better facial expression model than the full GaMo. For the less representative facial expressions like Disgust, Fear and Sad, the improvement is huge. The reason for this is that with equal consideration of all facial expressions during the training process, all expressions' deep features can be learned correctly, and if the test data can be well represented by the training data, we can achieve very good results. So, with our framework, we have a better chance to be able to obtain a robust expression predictor on all facial categories. 
\subsection{Comparison in  user feedback }

\begin{figure}
  \includegraphics[width=0.5\textwidth] {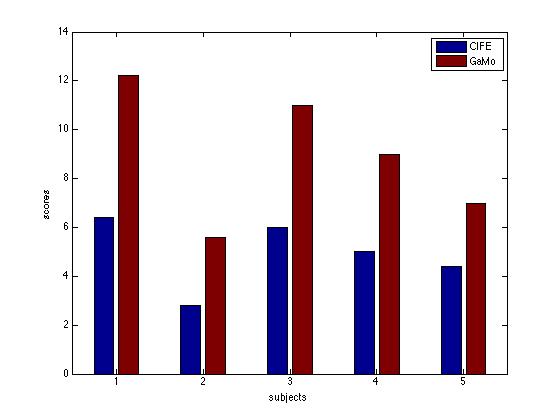}
\caption{Users' average scores on two GaMo and CIFE based CNN models}
\label{fig8}       
\end{figure}

The goal of facial expression recognition research is often to train a model that can perform well in real scenes. This is especially true in human-computer interaction  applications for real daily activities, such as satisfaction studies of customers and viewers, and assistive social interaction for people in need, or individuals with visual impairment and autism spectrum disorders (ASD). One approach to verify an expression detector is through a test on ordinary people with natural facial expressions. To accurately evaluate the two models, we analyze the data collected from five new users  (3 male and 2 female) who are not included in the GaMo dataset, while playing the general version of the game. Note that in the phase of GaMo data collection, we mainly use the customized game interface since users cannot perform well with the general game interface. In this game engine performance study, the general game is played five times by each user with the same game settings and the scores  of the five rounds are recorded. Using the two versions of our game engine, one trained on CIFE and the other on GaMo, respectively. Figure \ref{fig8} shows the result of this experiment. We have plotted the two average scores for each player on games powered by the two game engines. According to this figure, the GaMo game engine has a much better performance and results in higher scores. This further confirms that the model trained on GaMo is more suitable for real-world expression recognition.

This  result agrees with the cross testing results which show that the GaMo model has a better performance on the GaMo dataset itself. These observations would also support our claim that GaMo is very useful in detecting subtle expressions. For instance, the user can gain a point with a normal smile expression in GaMo model game as shown in Figure \ref{fig7}, while using the CIFE model, the expression can not be detected. Same fact holds for detecting anger or any other expressions, as our players do not have any prior knowledge of how obvious and explicit their facial expression should look like.

\section{Conclusion and Discussion}
In this paper, we propose a recursive framework  in order to achieve real scene facial expression recognition. We first build a  candid image facial expression dataset CIFE by parsing Web expression images from image search engines. The CNN-based deep learning approaches are then employed to train robust facial expression predictors, while fine-tuning approaches are also constructed to improve facial expression accuracy. To collect real scene images, we have designed a facial expression interaction game based  on our deep learning model that was trained with the CIFE dataset. With users playing both the general and the customized versions of the face game, the correctly labeled facial emotion images are selected and saved, which help us build the GaMo dataset. To prove the effectiveness of our framework, we compared GaMo and CIFE for their balanceness, recognition accuracy, the effectiveness of using strictly balanced subsets, and feedback from human subject tests. The experiments show that our framework can help build a reliable facial expression predictor for real scenes.  

Through our evaluation of the GaMo and CIFE datasets, we see the effectiveness of our framework. By recursively updating our model with newly collected data, we can achieve better facial expression recognition model for real scenes. By comparing the statistics of CIFE and GaMo trained models, we can obtain a more balanced GaMo dataset than the CIFE dataset. We can have more "under-represented" facial expressions using the game interface than collecting them from image search engines with tremendous manual efforts. We also have pointed out that we can use the  known recognition rate of our game engine for each emotion category to change the appearing frequency of them so we can obtain more balanced samples across the seven expressions.  The testing results on GaMo datasets, the real scene images, hold the fact that the models trained with GaMo have better performance in real scenarios. In the balanced subset experiment, GaMo shows us the potential to build a robust model able to detect all expressions. And finally in our human subject experiment, we saw the ability of our updated model to detect more subtle expressions. This leads us to believe that with more game data and by recursively updating our facial expression models, we can detect facial expressions in real scenes with better accuracy.



\begin{acknowledgements}

The first author would also like to thank IBM China Research Lab for the summer internship that enables the collection of the CIFE dateset. Special thanks to Ms Celina M. Cavalluzzi, Director of Day Services, GoodWill, for her assistance in evaluating the game apps by adults with Autism Spectrum Disorders. 
\end{acknowledgements}



\end{document}